\documentclass{article}
\usepackage{ijcai19}

\usepackage{times}
\usepackage{soul}
\usepackage[utf8]{inputenc}
\usepackage[small]{caption}
\usepackage{graphicx}
\usepackage{amsmath}
\usepackage{booktabs}
\usepackage{algorithm}
\usepackage{algorithmic}
\usepackage{natbib}
\usepackage{multibib}
\usepackage{subcaption}
\usepackage{verbatim}
\usepackage[export]{adjustbox}
\usepackage[hidelinks]{hyperref}
\newcommand{\beginsupplement}{%
        \setcounter{table}{0}
        \renewcommand{\thetable}{S\arabic{table}}%
        \setcounter{figure}{0}
        \renewcommand{\thefigure}{S\arabic{figure}}%
        \setcounter{section}{0}
        \renewcommand{\thesection}{S\arabic{section}}%
     }
     
\newif\ifcomments

\commentstrue

\ifcomments
\newcommand{\comments}[1]{#1}
\else
\newcommand{\comments}[1]{}
\fi

\usepackage{latexsym} 
\usepackage{EvolvingAIcommenting}

\title{An Atari Model Zoo for Analyzing, Visualizing, and Comparing Deep Reinforcement Learning Agents}

\author{
Felipe Petroski Such$^1$ \and  Vashisht Madhavan$^1$ \and Rosanne Liu$^1$ \and Rui Wang$^1$ \\ Pablo Samuel Castro$^2$  \and
Yulun Li$^1$ \and Jiale Zhi$^1$ \and Ludwig Schubert$^3$ \\  Marc G. Bellemare$^2$ Jeff Clune$^1$  \and  Joel Lehman$^1$\thanks{Corresponding author: \texttt{joel.lehman@uber.com}} \\
  $^1$Uber AI Labs\and$^2$Google Brain\and$^3$OpenAI\\
}

\newcites{New}{References}
\begin{document}

\maketitle

\begin{abstract}
Much human and computational effort has aimed to improve
how deep reinforcement learning (DRL) algorithms perform on benchmarks such as the Atari Learning Environment. Comparatively less effort has focused on understanding what has been learned by such methods, and investigating and comparing the representations learned by different families of DRL algorithms. Sources of friction include the onerous computational requirements, and general logistical and architectural complications for running DRL algorithms at scale. We lessen this friction, by (1) training several algorithms at scale and releasing trained models, (2) integrating with a previous DRL model release, and (3) releasing code that makes it easy for anyone to load, visualize, and analyze such models. This paper introduces the Atari Zoo framework, which contains models trained across benchmark Atari games, in an easy-to-use format, as well as code that implements common modes of analysis and connects such models to a popular neural network visualization library. Further, to demonstrate the potential of this dataset and software package, we show initial quantitative and qualitative comparisons between the performance and representations of several DRL algorithms, highlighting interesting and previously unknown distinctions between them.
\end{abstract}

\section{Introduction}

Since its introduction the Atari Learning Environment (ALE; \citep{bellemare:ale}) has 
been an important reinforcement learning (RL) testbed. It enables easily evaluating 
algorithms on over 50 emulated Atari games spanning diverse gameplay styles, providing
a window on such algorithms' generality. Indeed, surprisingly strong results in ALE with
deep neural networks (DNNs), published in Nature \citep{mnih:human}, greatly
contributed to the current popularity of deep reinforcement learning (DRL).

Like other machine learning benchmarks, much effort aims to quantitatively improve
state-of-the-art (SOTA) scores. As the DRL community grows, a paper pushing SOTA is likely to
attract significant interest and accumulate citations. While improving performance is important, it is equally important to understand what DRL algorithms learn, how they process and represent information, and what are their properties, strengths, and weaknesses. These questions cannot be answered through simple quantitative measurements of performance across the ALE suite of games.


Compared to pushing SOTA, much less 
work has focused on understanding, interpreting, and visualizing \emph{products} of DRL; in particular,
little research compares DRL algorithms across dimensions other than performance.
This paper thus aims to alleviate the considerable \emph{friction} for those looking to rigorously understand the qualitative behavior of DRL agents. 
Three main sources of such friction are: (1) the significant computational
resources required to run DRL at scale, (2) the logistical tedium of plumbing the products of different DRL algorithms into
a common interface, and (3) the wasted effort in re-implementing standard analysis pipelines (like t-SNE embeddings of the state space \citep{mnih:human}, 
or activation maximization for visualizing what neurons in a model represent \citep{erhan2009visualizing,olah:buildingblocks,nguyen2017plug,simonyan2013deep,yosinski:understanding,mahendran2016visualizing}).
To address these frictions, this paper introduces
the \emph{Atari Zoo}, a release of trained models spanning major families of DRL algorithms, and an accompanying 
open-source software package\footnote[1]{\url{https://github.com/uber-research/atari-model-zoo}}
that enables their easy analysis, comparison, and visualization (and similar analysis of future models). In particular, this package
enables easily downloading particular frozen models of interest from the zoo on-demand, further evaluating them in 
their training environment or modified environments, generating visualizations of their neural activity, exploring 
compressed visual representations of their behavior, and creating synthetic input patterns that reveal what particular
neurons most respond to.


To demonstrate the promise of this model zoo and software, this paper presents an initial analysis of the products of seven
DRL algorithms spanning policy gradient, value-based, and evolutionary methods\footnote[2]{While evolutionary algorithms are excluded from some definitions of RL, their inclusion in the zoo can help investigate what distinguishes such black-box optimization from more traditional RL.}: A2C (policy-gradient; \citep{mnih2016asynchronous}), IMPALA (policy-gradient; \citep{espeholt:impala}), DQN (value-based; \citep{mnih:human}), Rainbow (value-based; \citep{hessel2017rainbow}), Ape-X (value-based; \citep{horgan:distributed}), ES (evolutionary; \citep{salimans:es}), and Deep GA (evolutionary; \citep{such:deep}). The analysis illuminates differences in learned policies across methods that are independent of raw score performance, highlighting the benefit of 
going beyond simple quantitative measures and of having a unifying software framework that enables analyses with multiple, different, complementary techniques and applying them across many RL algorithms.



\vspace{-0.1in}
\section{Background}


\subsection{Visualizing Deep Networks}
One line of DNN research focuses on visualizing the internal dynamics of a DNN \citep{yosinski:understanding} or
examines what particular neurons detect or respond to \citep{erhan2009visualizing, zeiler:visualizing,olah:buildingblocks,nguyen2017plug,simonyan2013deep,yosinski:understanding,mahendran2016visualizing}. The hope is
to gain more insight into how DNNs are representing information, motivated both to enable more interpretable
decisions from these models \citep{olah:buildingblocks} and to illuminate previously unknown properties about DNNs \citep{yosinski:understanding}. 
For example, through live visualization of all activations in a vision network responding to different images, \citet{yosinski:understanding} 
highlighted that representations were often surprisingly local (as opposed to distributed), e.g.\ one convolutional filter proved to be a reliable face
detector. One practical value of such insights is that they can catalyze future research.
The Atari Zoo enables animations in the spirit of
\citet{yosinski:understanding} that show an agent's activations as it interacts with a game, and also enables creating synthetic inputs via activation maximization \citep{erhan2009visualizing, zeiler:visualizing,olah:buildingblocks,nguyen2017plug,simonyan2013deep,yosinski:understanding,mahendran2016visualizing}, specifically by connecting DRL agents to the Lucid visualization package \citep{Lucid}. 

\subsection{Understanding Deep RL}
While much more visualization and understanding work has been done for vision models than for DRL, a few papers directly focus on understanding DRL agents \citep{greydanus:visualizing,zahavy:dqn}, 
and many others feature some analysis of DRL agent behavior (often in the form of t-SNE diagrams of the state space; see \citep{mnih:human}).
One approach to understanding DRL agents is to investigate the learned features of models \citep{greydanus:visualizing,zahavy:dqn}. For example, \citet{zahavy:dqn} visualize what pixels are most
important to an agent's decision by using gradients of decisions with respect to pixels.
Another approach is to modify the DNN architecture or training procedure such that a trained model will have more interpretable features 
\citep{annasamy2018towards}. For example, \citet{annasamy2018towards} 
augment a model with an attention mechanism and a reconstruction loss, hoping to produce interpretable explanations as
a result. 

The software package released here is in the spirit of the first paradigm. It facilitates understanding the most commonly applied architectures instead of changing them, although it is designed also to accommodate importing in new vision-based DRL models,
and could thus be also used to
analyze agents explicitly engineered to be interpretable. In particular, the package enables re-exploring many past DRL analysis techniques at scale, and across algorithms, which were previously applied only for one algorithm and across only
a handful of hand-selected games.

\subsection{Model Zoos}

A useful mechanism for reducing friction for analyzing and building upon models is the idea of a \emph{model zoo}, i.e.\ a repository of
pre-trained models that can easily be further investigated, fine-tuned, and/or compared (e.g.\ by looking at how their high-level representations differ). For example, the Caffe website includes a model zoo with many popular vision models, as do Tensorflow, Keras, and PyTorch.
The idea is that training large-scale vision networks (e.g.\ on the ImageNet dataset) can take weeks with powerful GPUs, and that there is little reason to constantly reduplicate the
effort of training. Pre-trained word-embedding models are often released with similar motivation, e.g.\ for Word2Vec or GLoVE. 
However, such a practice is much
less common in the space of DRL; one reason is that so far, unlike with vision models and word-embedding models, there are few other down-stream tasks from which Atari DRL agents provide obvious value. But, if the goal is to better understand these models and algorithm, both to improve them and to use them safely, then there is value in their release.

The recent Dopamine reproducible DRL package \citep{bellemare:dopamine} released trained ALE models; it includes final checkpoints of models trained by several DQN variants. However,
in general it is non-trivial to extract TensorFlow models from their original context for visualization purposes, and to compare agent behavior across DRL algorithms in the same software framework (e.g.\ due to slight differences in image preprocessing), or to explore dynamics that take place \emph{over} learning, i.e.\ from \emph{intermediate} checkpoints. To remedy this, for this paper's accompanying software release, the Dopamine checkpoints were distilled into frozen models that can be easily loaded into the Atari Zoo framework; and for algorithms trained specifically for the Atari Zoo, we distill intermediate checkpoints in addition to final ones.

\section{Generating the Zoo}

The approach is to run several validated implementations of DRL algorithms and to collect and standardize the models and results, such that they can then be easily used for downstream analysis and synthesis. There are many algorithms, implementations of them, and different ways that they could be run (e.g.\ different hyperparameters, architectures, input representations, stopping criteria, etc.). These choices influence the kind of post-hoc analysis that is possible. For example, Rainbow most often outperforms DQN, and if only final models are released, it is impossible to explore
scientific questions where it is important to \emph{control} for performance.

We thus
adopted the high level principles that the Atari Zoo should hold as many elements of architecture and experimental design constant across algorithms (e.g.\ DNN structure, input representation), should enable as many types of downstream analysis as possible (e.g.\ by releasing checkpoints across training time), and should make reasonable allowances for the particularities of each algorithm (e.g.\ ensuring hyperparameters are well-fit to the algorithm, and allowing for differences in how policies are encoded or sampled from). The next paragraphs describe specific design choices. 

\subsection{Frozen Model Selection Criteria}

To enable the platform to facilitate a variety of explorations, we release multiple frozen models for each run, according to different criteria that may be useful to control for when comparing learned policies. The idea is that depending on the desired analysis, controlling for samples, or for wall-clock, or for performance (i.e.\ comparing similarly-performing policies) will be more appropriate. In particular, in addition to releasing the final model for each run, additional are models taken over training time (at one, two, four, six, and ten hours); over game frame samples (400 million, and 1 billion frames); over scores (if an algorithm reaches human level performance); and also a model before any training, to enable analysis of how weights change from their random initialization. The hope is that these frozen models will cover a wide spectrum of possible use cases.


\subsection{Algorithm Choice}

One important choice for the Atari Zoo is which DRL algorithms to run and include. The main families of DRL algorithms that have been applied to the ALE are policy gradients methods like A2C \citep{mnih2016asynchronous}, value-based methods
like DQN \citep{mnih:human}, and black-box optimization methods like ES \citep{salimans:es} and Deep GA \citep{such:deep}. Based on representativeness and 
available trusted implementations, 
the particular algorithms chosen to train included two policy gradients algorithms (A2C \citep{mnih2016asynchronous} and IMPALA \citep{espeholt:impala}), two evolutionary algorithms (ES \citep{salimans:es} and Deep GA \citep{such:deep}), and one value-function based algorithm (a high-performing DQN variant, Ape-X; \citep{horgan:distributed}). Additionally, models are also
imported from the Dopamine release \citep{bellemare:dopamine}, which include DQN \citep{mnih:human} and a sophisticated variant of it called Rainbow \citep{hessel2017rainbow}. Note that from the Dopamine models, only final models are currently available. Hyperparameters and training details for all algorithms are available in supplemental material section \ref{sec:algo_hyper}. We hope to include models from additional algorithms in future releases.

\subsection{DNN Architecture and Input Representation}

All algorithms are run with the DNN architecture from \citet{mnih:human}, which consists of three convolutional layers (with filter size 8x8, 4x4, and 3x3, followed by a fully-connected layer). For most of the explored algorithms, the fully-connected layer connects to an output layer with one neuron per valid action in the underlying Atari game. However, A2C and IMPALA have an additional output that approximates the state value function; Ape-X's architecture features dueling DQN \citep{wang2015dueling}, 
which has two separate fully-connected streams; and Rainbow's architecture includes C51 \citep{bellemare2017distributional}, 
which uses many outputs to approximate the distribution of expected Q-values.

\begin{figure}[t]
\begin{center}
\begin{subfigure}{.13\textwidth}
  \centering
	\vspace{0.35in}
  \includegraphics[width=.85\linewidth]{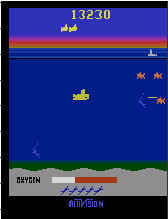}
	\vspace{0.1in}
  \caption{RGB frame}
\end{subfigure}
\begin{subfigure}{.19\textwidth}
  \centering
  \includegraphics[width=1.0\linewidth]{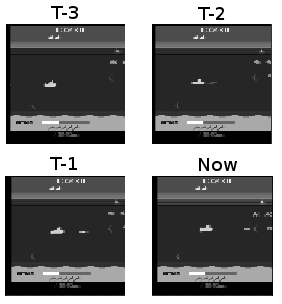}
  \caption{Observation}
\end{subfigure}
\begin{subfigure}{.14\textwidth}

  \vspace{0.2in}
  \centering
  \includegraphics[width=.6\linewidth]{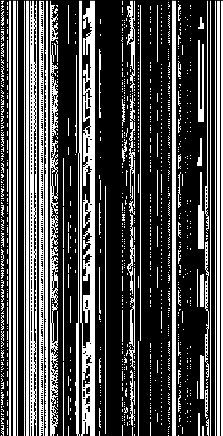}
  \vspace{0.1in}
  \caption{RAM state}
\end{subfigure}
	\vspace{-0.05in}
\caption{\textbf{Input and RAM Representation}. (a) One RGB frame of emulated Atari gameplay is shown, which is (b) preprocessed and concatenated with previous frames before being fed as an observation into the DNN agent. A compressed representation of a 2000-step ALE simulation is shown in (c), i.e.\ the 1024-bit RAM state (horizontal axis) unfurled over frames (vertical axis).}
\label{fig:representation}
\end{center}
	\vspace{-0.2in}
\end{figure}

Atari frames are 210x160 color images (see figure \ref{fig:representation}a); the canonical DRL representation is a a tensor consisting of the four most recent observation frames, grayscaled and downsampled to 84x84 (figure \ref{fig:representation}b). By including some previous frames, the aim is to make the game more fully-observable, to boost performance of the feed-forward architectures that are currently most common in Atari research (although recurrent architectures offer possible improvements \citep{mnih2016asynchronous,espeholt:impala}).
One useful Atari representation that is applied in post-training analysis in this paper, is the Atari RAM state, which is only 1024 bits long but encompasses the true underlying state (figure \ref{fig:representation}c).

\subsection{Data Collection}

All algorithms are run across 55 Atari games, for at least three independent random weight initializations. Regular checkpoints were taken during training; after training, the checkpoints that best fit each of the desired criteria (e.g.\ 400 million frames or human-level performance) were frozen and included in the zoo. The advantage of this post-hoc culling is that additional criteria can be added in the future, e.g.\ if Atari Zoo users introduce a new use case, because the original checkpoints are archived. Log files were stored and converted into a common format that are also released with the models, to aid future performance curve comparisons for other researchers. Each frozen model was run post-hoc in ALE for $2500$ timesteps to generate cached behavior of policies in their training environment, which includes the raw game frames, the processed four-frame observations, RAM states, and high-level representations (e.g.\ neural representations at hidden layers). As a result, it is possible to do meaningful analysis without ever running the models themselves.






\section{Quantitative Analysis}

The open-source software package released with the acceptance of this work 
provides an interface to the Atari Zoo dataset, and implements several
common modes of analysis. Models can be downloaded with a single line of code; and other single-line invocations  interface directly with ALE and return the behavioral outcome of executing a model's policy, or create movies of agents superimposed with neural activation, or access convolutional weight tensors. In this section, we demonstrate analyses the Atari Zoo software can facilitate, and highlight some of its built-in features. For many of the analyses below, for computational simplicity we study results in a representative subset of 13 ALE games used by prior research \citep{such:deep}, which we refer to here as the \emph{analysis subset} of games.

\subsection{Convolutional Filter Analysis}
\label{sec:conv_analysis}

While understanding a DNN only by examining its weights is challenging, weights directly connected to the input can often be interpreted. For example, from visualizing the weights of the first convolutional layer in a vision model, Gabor-like edge detection filters are nearly always present. 
An interesting question is if Gabor-like filters also arise when DRL algorithms are trained from pixel input (as is done here). In visualizing filters across games and DRL algorithms, edge-detector-like features sometimes arise in the gradient-based methods, but they are seemingly never as crisp as in vision models; this may because ALE lacks the visual complexity of natural images. In contrast, the filters in the evolutionary models are less regular. Representative examples across games and algorithms are shown in supplemental figure \ref{fig:convfilters}. 

Learned filters commonly are tiled similarly across time (i.e.\ across the four DNN input frames), with past frames having lower-intensity weights. One explanation is that reward gradients are more strongly influenced by present observations. To explore this systematically, across games and algorithms we examined the absolute magnitude of filter weights connected to the present frame versus the past. In contrast to the gradient-based methods the evolutionary methods show no discernable preference across time (supplemental figure \ref{fig:temporal}), again suggesting that their learning differs qualitatively from the gradient-based methods. 
Interestingly, a rigorous information-theoretic approximation of memory usage is explored by \citet{dann2016memory} in the context of DQN; our measure well-correlates with theirs despite the relative simplicity of exploring only filter weight strength (supplemental section \ref{sec:memory}).  

\subsection{Robustness to Observation Noise}
An important property is how agents perform in slightly out-of-distribution (OOD) situations; ideally they would not catastrophically fail in the face of nominal change. While it is difficult to 
freely alter the ALE game dynamics (without learning how to program in 6502 assembly code), it is possible to systematically distort observations. Here we
explore one simple OOD change to observations by adding increasingly severe noise to the observations input to DNN-based agents, and observe how their evaluated game score degrades. The motivation is to discover whether some learning algorithms are learning more robust policies than others. The results show that with some caveats, methods with a direct representation of the policy appear more robust to observation noise (supplemental figure \ref{fig:obs_noise}).
A similar study conducted for robustness to parameter noise (supplemental section \ref{sec:parameter_noise}) tentatively suggests that actor-critic methods are more robust to such noise.

\subsection{Distinctiveness of Learned Policies}



To explore the distinctive signature of solutions discovered by different DRL algorithms, we train image classifiers to identify the generating DRL algorithm given states sampled from independent runs of each algorithm (details are in supplemental section \ref{sec:state}). 
Supplemental figure \ref{fig:classify_cf} shows the confusion matrix for Seaquest, wherein a cluster of policy search methods (A2C, ES, and GA) have the most inter-class confusion, reflecting (as confirmed in later sections) that these algorithms tend to converge to the same sub-optimal behavior in this game; results are qualitatively similar when tabulated across the analysis subset of games (supplemental figure \ref{fig:classify_cf_all}). 

\begin{figure}
\centering
\includegraphics[width=.9\linewidth]{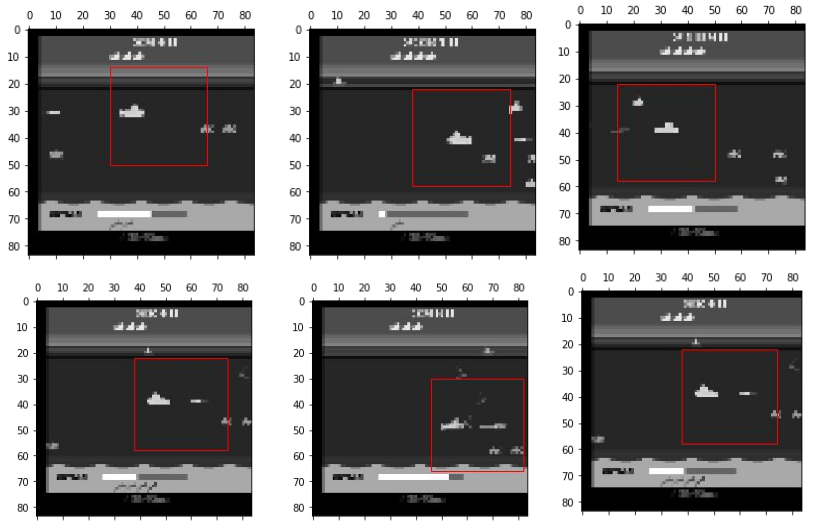}
	\vspace{-0.075in}
	\caption{\textbf{A sub-detecting neuron in Seaquest}. Each image represents an observation from an Ape-X agent
	playing Seaquest. The red square indicates which image patch highly-activated the sub-detecting neuron on the
	third convolutional layer of the DNN. Having multiple tools (such as this image patch finder, or the activation 
	movies which identified this neuron of interest) enables more easily triangulating and verifying hypotheses about
	the internals of a DRL agent's neural network.}
\label{fig:subdetector}
\end{figure}

\begin{figure*}[ht]
\centering
\includegraphics[width=.56\linewidth]{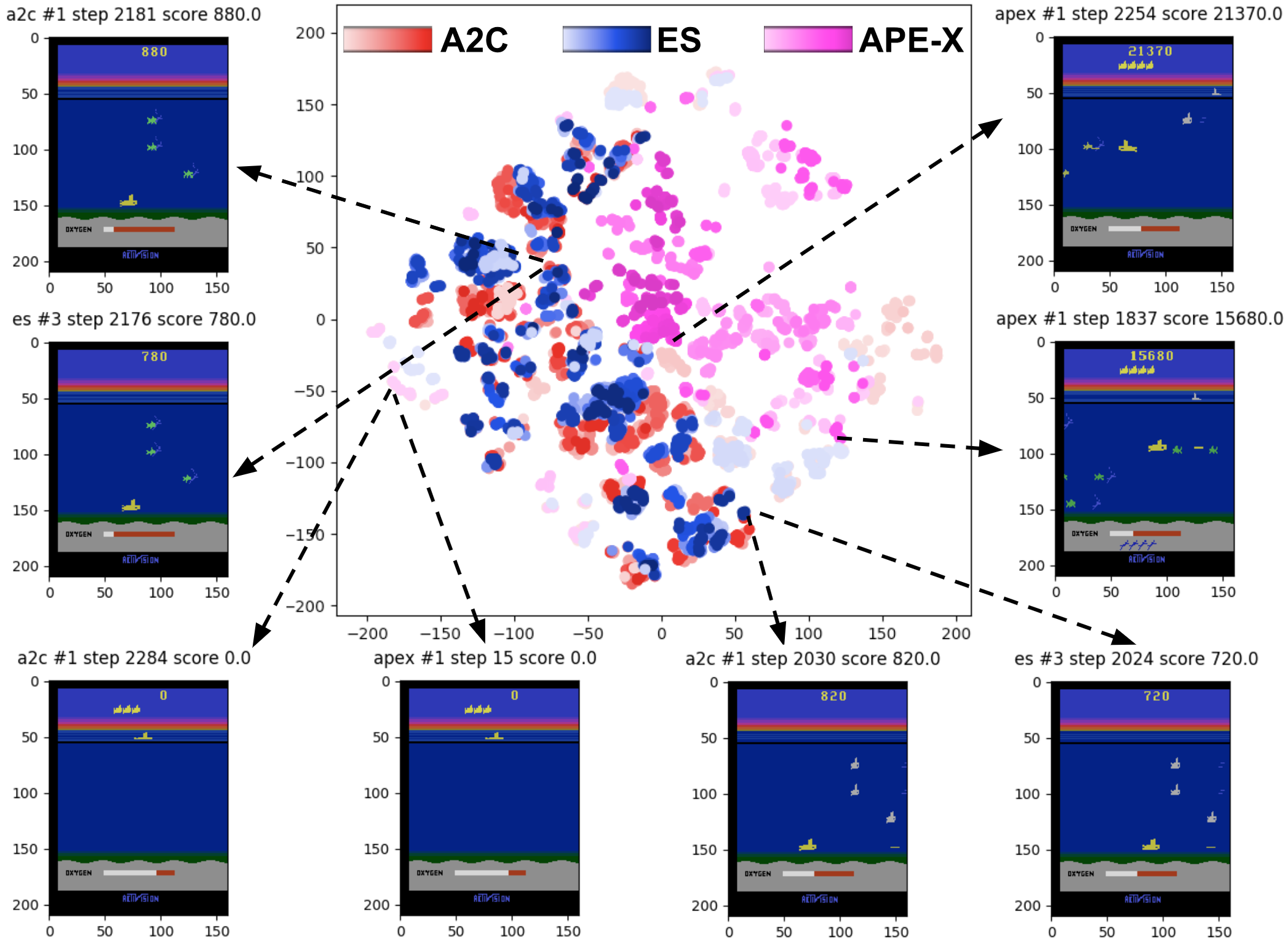}
	\vspace{-0.1in}
\caption{\textbf{Multiple runs of algorithms and sharing the same RAM-space embedding in Seaquest}. This plot shows one ALE evaluation per model for A2C, ES, and Ape-X, visualized in the
same underlying RAM t-SNE embedding. Each dot represents a separate frame from each agent, colored by score (darker color indicates higher score). The plot highlights that in this game, A2C and ES visit similar distributions of states (corresponding to the same sub-optimal behavior), while Ape-X visits a distinct part of the state-space, i.e.\ matching what could manually be distilled from watching the policy movies shown in supplemental figure \ref{fig:grid_rollout}. The interface allows clicking on points to observe the corresponding RGB frame, and for toggling different runs of different algorithms for visualization.}
\label{fig:tsne_RAM}
\end{figure*}

\vspace{-0.1in}
\section{Visualization}

We next highlight the Atari Zoo's capabilities to quickly
and systematically visualize policies, which
broadly can be divided into three categories: Direct policy visualization, dimensionality reduction, and
neuron activation maximization. 

\subsection{Animations to Inspect Policy and Activations}
\label{sec:act_video}

To quickly survey the solutions being learned, our software generates grids of videos, where
one grid axis spans different DRL algorithms, and the other axis covers independent runs of
the algorithm. Such videos can highlight when different algorithms are converging to the same
local optimum (e.g.\ supplemental figure \ref{fig:grid_rollout} shows a situation where this is the case for
A2C, ES, and the GA; video: \url{http://bit.ly/2XpD5kO}).

\begin{figure}
\centering
\includegraphics[width=.6\linewidth]{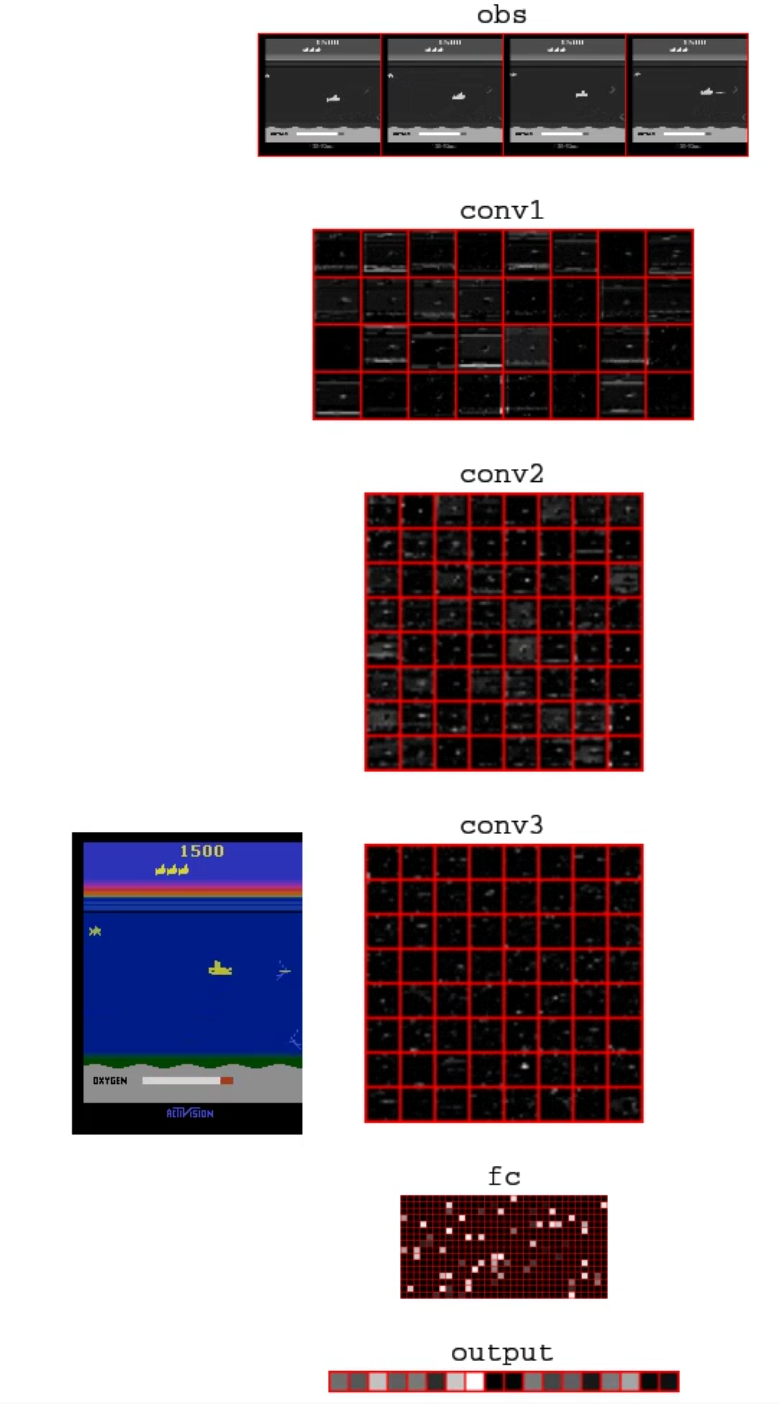}
	\caption{\textbf{Policy and activation visualization}. The figure shows a still frame from a video of an Ape-X agent acting in the Seaquest environment (full video can be accessed at 
	\url{http://bit.ly/2tFHiCU}).
	On the left, the RGB frame is shown, while from top to bottom on the right are: the processed observations, and then the activations for the convolutional layers, the fully connected layer, and finally, the Q-value outputs. From watching the video, it is apparent that the brightest neuron in the third convolutional layer tracks the position of the submarine. This shows that like in vision DNNs, sometimes important features are represented in a local, rather than distributed fashion \protect\citepNew{yosinski:understanding}.}
\label{fig:activation_movie}
\end{figure}

To enable investigating the internal workings of the DNN, our software generates movies 
that display activations of all neurons alongside animated frames of the agent acting in
game. This approach is inspired by the deep visualization toolbox \citep{yosinski:understanding}, but
put into a DRL context. Supplemental figure \ref{fig:activation_movie} shows how this tool can lead to recognizing
the functionality of particular high-level features (video: \url{http://bit.ly/2tFHiCU}); 
in particular, it helped to identify a
submarine detecting neuron on the third convolution layer of an Ape-X agent. Note that
for ES and GA, no such specialized neuron was found; activations seemed qualitatively
more distributed for those methods.

\subsection{Image Patches that Maximally Excite Filters}
\label{sec:patches}

One automated technique for uncovering the functionality of a particular convolutional filter is to find which image patches evoke from it the highest magnitude activations. 
Given a trained DRL agent and a target convolution filter to analyze, observations from the agent interacting with its ALE training environment are input to the agent's DNN, and resulting maps of activations from the filter
of interest are stored. These maps are sorted by the single maximum activation within them, and the geometric location within the map of that maximum activation is recorded. 
Then, for each of these top-most activations, the specific image patch from the observation that generated it is identified and displayed, by taking the receptive field of the filter into account (i.e.\ modulated by
both the stride and size of the convolutional layers). As a sanity check, we validate that the neuron identified in the previous section does indeed maximally fire for submarines (figure \ref{fig:subdetector}).

\begin{figure}[ht]
\vspace{-0.2in}
  \centering
	\includegraphics[width=0.9\linewidth,trim={0in 0in 10in 0in},clip]{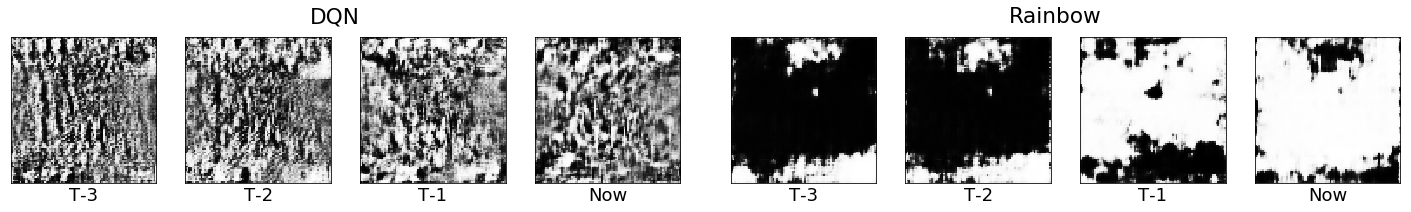}
	\includegraphics[width=0.9\linewidth,trim={10.0in 0in 0in 0in},clip]{figures/new/seaquest_out.png}
	\vspace{-0.1in}
  \caption{\textbf{Synthesized inputs for output layer neurons in Seaquest.} For a representative run of Rainbow and DQN, inputs are shown optimized to maximize the activation of the first neuron in the output layer of a Seaquest network. Because Rainbow includes C51, its image is in effect optimized to maximize the probability of a low-reward scenario; this neuron appears to be learning interpretable features such as submarine location and the seabed. When maximizing (or minimizing) DQN Q-value outputs (one example shown on left), this qualitative outcome of interpretability was not observed.}
  \label{fig:seaquest_final_layer}
\end{figure}

\subsection{Dimensionality Reduction}

\begin{figure}
\centering
\includegraphics[width=.75\linewidth]{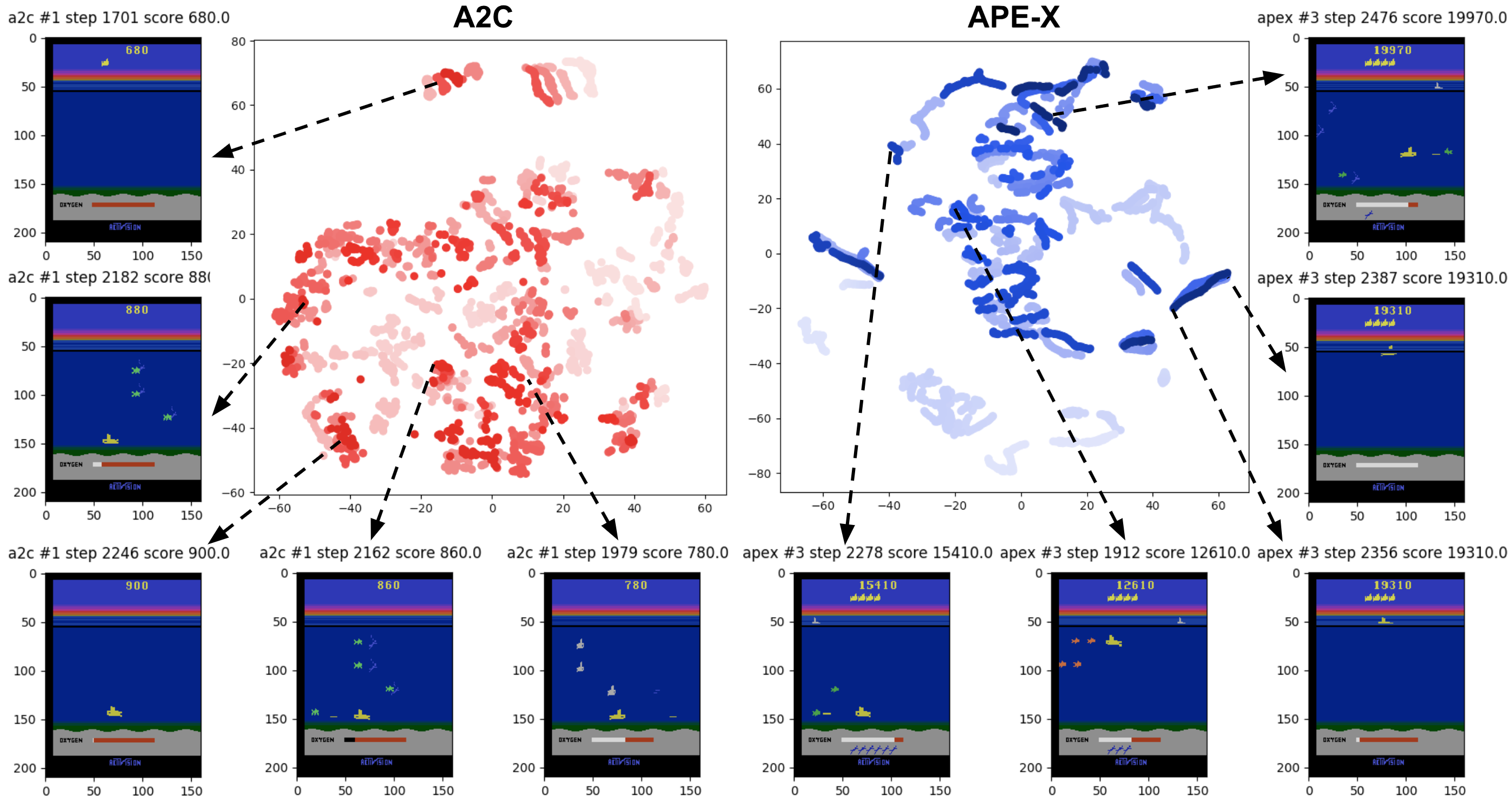}
\caption{\textbf{Comparing high-level DNN representations through separate t-SNE embeddings}. The figure shows separate t-SNE embeddings of high-level representations for DNNs trained to play Seaquest by A2C and Ape-X. Each dot corresponds to a specific frame in a rollout, and darker shades indicate higher scores. Embeddings that represent similar frames cluster together, indicating states with different positions of the submarine, and objects of various numbers, categories and colors. Representative frames for selected clusters are displayed. For example, in the left figure (A2C), the top-left cluster represents terminated states, and the bottom-left cluster corresponds to the situation of oxygen depletion, while in the right figure (Ape-X), bottom-right cluster corresponds to a repeated series of actions that the agent takes to surface and refill its oxygen. }
\label{fig:tsne_HLR}
\end{figure}

Dimensionality reduction provides another view on agent behavior; often DRL research includes 
t-SNE plots 
of agent DNN representations
that summarize behavior in the domain \citep{mnih:human}. Our software includes such an implementation
(supplemental figure \ref{fig:tsne_HLR}).

However, such an approach relies on embedding the high-level
representation of \emph{one} agent; it is unclear how to apply it to create an embedding appropriate
for comparisons of different independent runs of the same algorithm, or runs from different
DRL algorithms. As an initial approach, we implement an embedding
based on the Atari RAM representation (which is the same across algorithms and runs, but distinct
between games). Like the grid view of agent behaviors and the state-distinguishing classifier,
this t-SNE tool provides high-level information from which to compare runs of or between different algorithms 
(figure \ref{fig:tsne_RAM}); details of this approach are provided in supplemental
section \ref{sec:sup_tsne}.

\begin{figure*}[h]
  \centering
  \includegraphics[width=.82\linewidth]{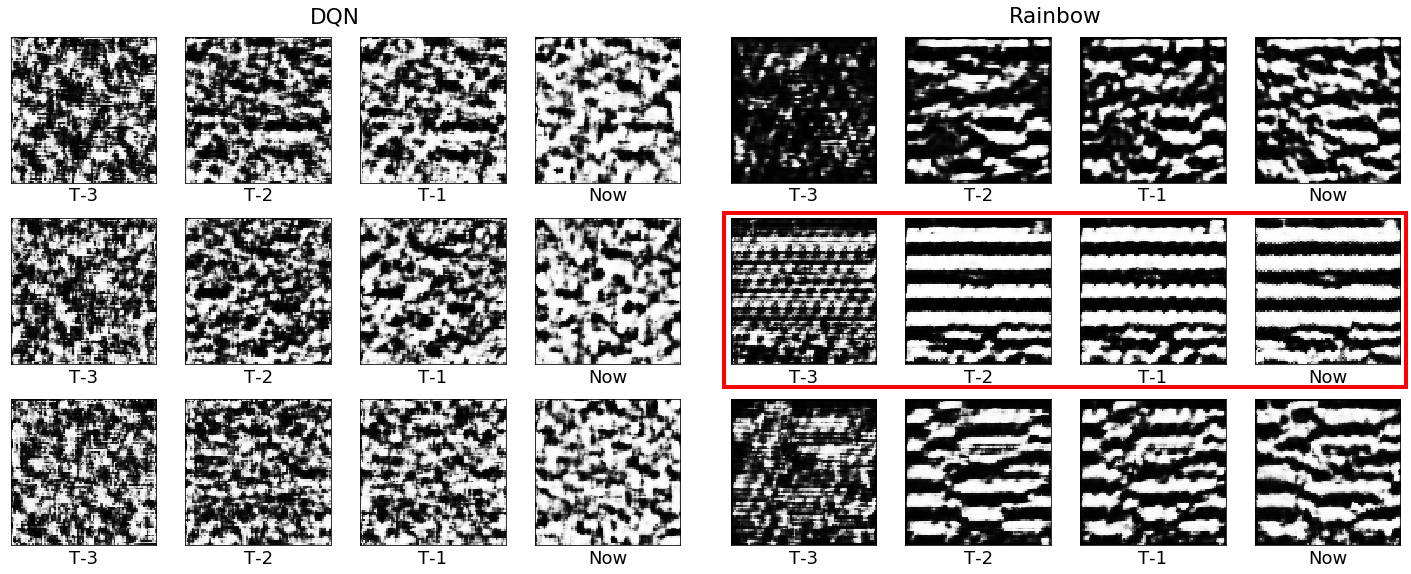}
	\vspace{-0.1in}
  \caption{\textbf{Synthesized inputs for fully-connected layer neurons in Freeway.} Inputs synthesized to maximize activations of the first three neurons in the first fully connected layer are shown for a respresentative DQN and  Rainbow DNN. One of the Rainbow neurons (in red rectangle) appears to be capturing lane features.}
  \label{fig:freeway_first_fc}
\end{figure*}

\vspace{-0.03in}
\subsection{Synthesizing Inputs to Understand Neurons}

While the previous sections explore DNN activations in the context of an agent's training environment, another approach is to optimize synthetic
input images that stimulate particular DNN neurons. Variations on this approach have yielded striking results in vision models \citep{nguyen2017plug,olah:buildingblocks,simonyan2013deep}; 
the hope is that these techniques could yield an additional view on 
DRL agents' neural representations. To enable this analysis, we leverage the Lucid visualization library \citep{Lucid}; 
in particular, we create wrapper classes that enable easy integration of Atari Zoo models into Lucid,
and release Jupyter notebooks that generate synthetic inputs for different DRL models. 

\begin{figure}
  \centering
  \includegraphics[width=.9\linewidth]{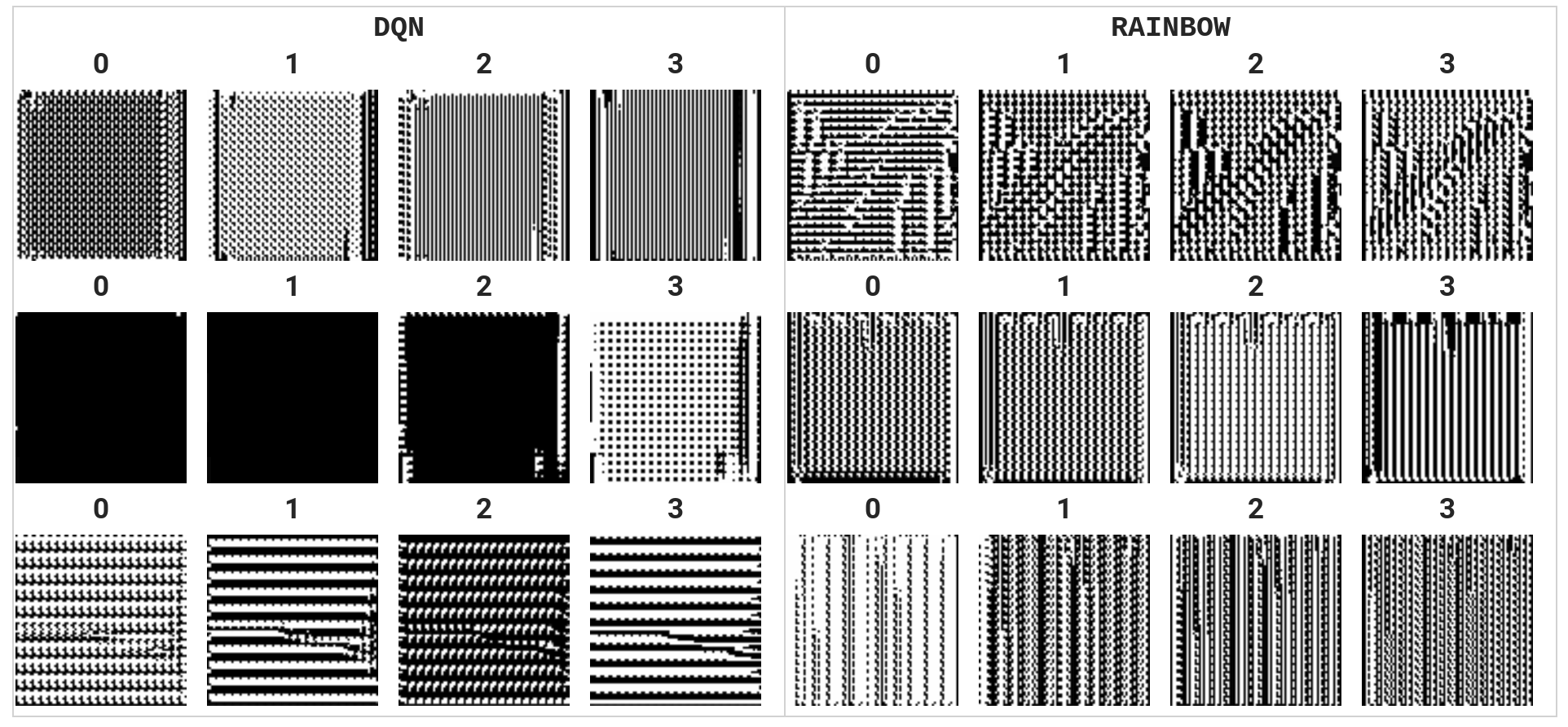}
	\caption{\textbf{Synthesized inputs for neurons in the first convolutional layer in Seaquest.} Inputs optimized to activate the first three neurons in the first convolutional layer are shown for representative runs of DQN and Rainbow. These neurons appear to be learning `edge-detector' style features.}
  \label{fig:seaquest_conv_relu}
\end{figure}

\begin{figure}
  \centering
  \includegraphics[width=.9\linewidth]{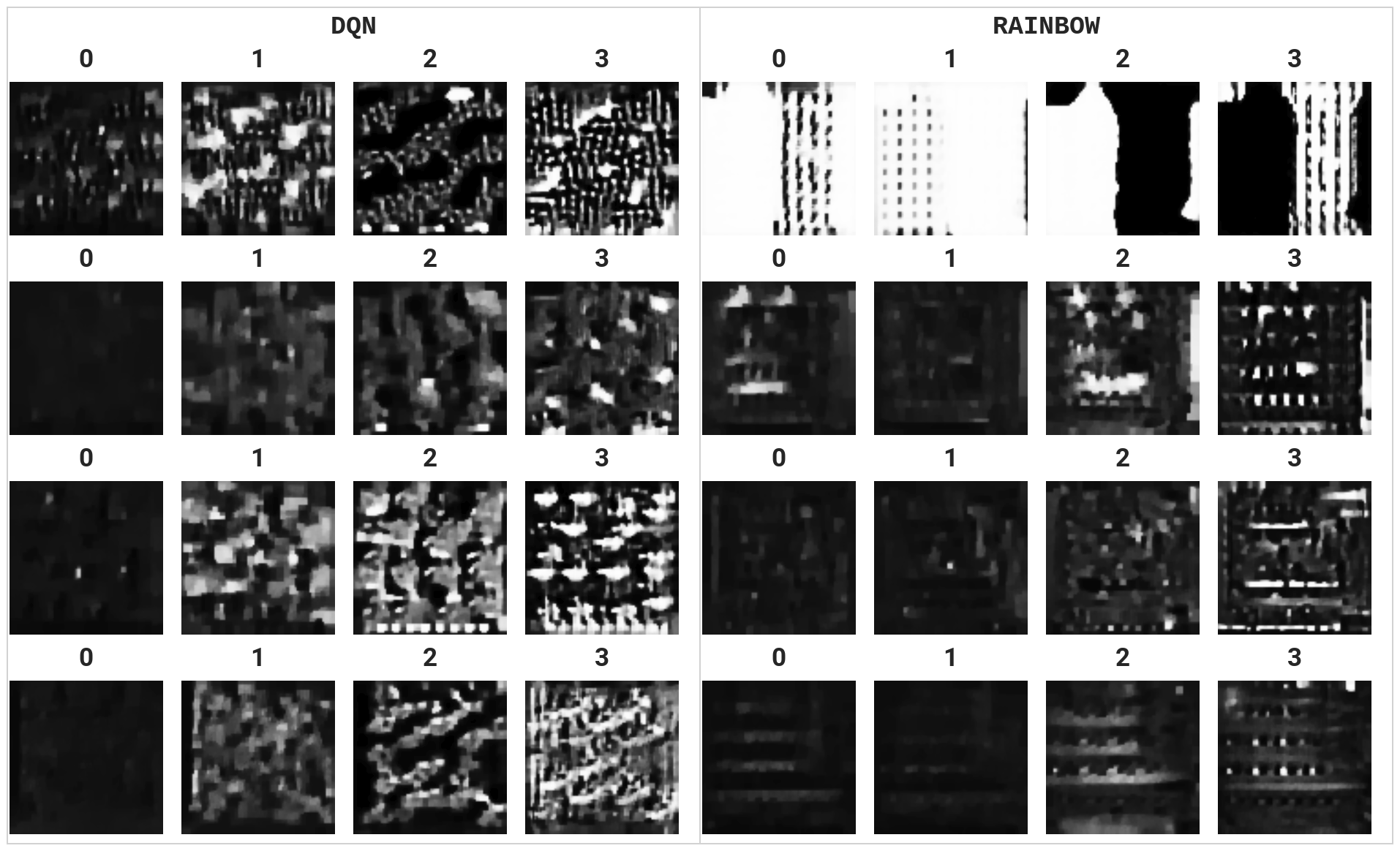}
	\caption{\textbf{Synthesized inputs for neurons in the third convolutional layer in Seaquest.} Inputs optimized to activate the first four neurons in the last (third) convolutional layer in Seaquest are shown for a representative run of DQN and Rainbow (hyperparameters for regularization, e.g.\ a total variation penalty, were optimized by hand to improve image quality). Both networks appear to focus on particular styles of objects, combinations of them, and animation-related features such as blinking. Some synthetic inputs make sense in the context of other investigatory tools; e.g.\ Rainbow's first neuron's synthetic input includes objects that blink between frames, and when explored with the patch activation technique, is seen responding most intensely when the sub is blinking and about to explode from running out of oxygen. 
	However, for some features it is unclear how the synthetic input is to be interpreted without further investigation, e.g.\ the patch activation technique shows that Rainbow's third neuron responds most when the sub is nearing the top border of the water.
Further experimentation with regularization within Lucid, or employing more sophisticated techniques, may help to improve these initial results.}
  \label{fig:seaquest_conv2_relu}
\end{figure}

We now present a series of synthetic inputs generated by the Lucid library across a handful of games that highlight the potential of these kinds of visualizations for 
DRL understanding (further details of the technique used are described in supplemental section \ref{sec:lucid}.
We first explore the kinds of
features learned across depth. Supplemental figure \ref{fig:seaquest_conv_relu} supports what was
learned by visualizing the first-layer filter weights for value-based networks (section \ref{sec:conv_analysis}; i.e.\ showing that first convolution layers in the value-based networks appear to be learning edge-detector features). The activation videos of section \ref{sec:act_video} and the patch-based approach of section \ref{sec:patches} help to provide grounding, showing that in the context of the game, some first-layer filters detect the edges of the screen, in effect to serve as location anchors, while others encode concepts like blinking objects (see figure \ref{fig:anchors}). Supplemental figure \ref{fig:seaquest_conv2_relu} explores visualizing later-layer convolution filters, and figure \ref{fig:seaquest_final_layer} show inputs synthesized to maximize output neurons, which sometimes yields interpretable features. 

\begin{figure}
  \centering
  \includegraphics[width=.9\linewidth]{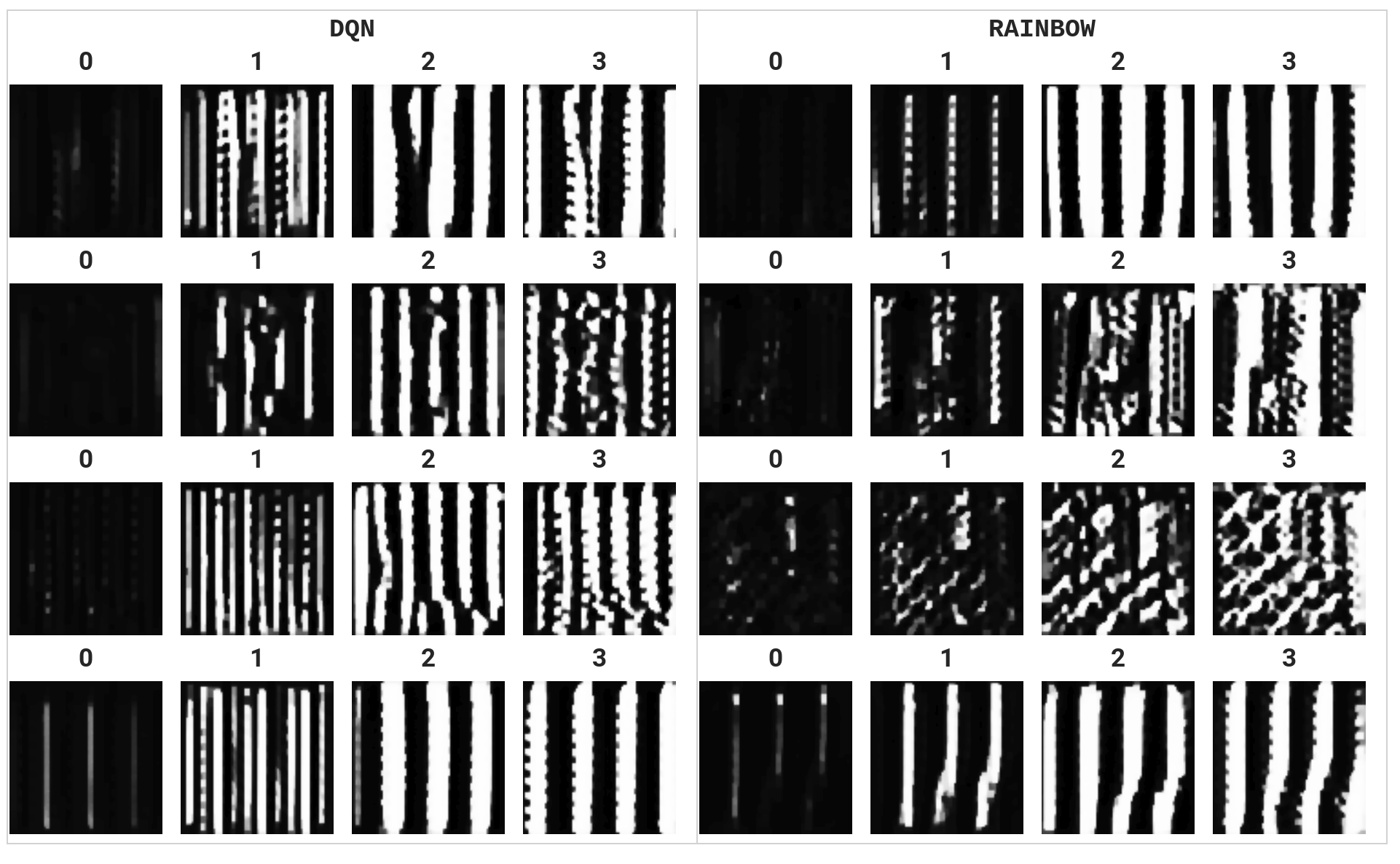}
	\caption{\textbf{Synthesized inputs for neurons in the third convolutional layer in Pong.} Inputs optimized to activate the first four neurons in the last (third) convolutional layer in Pong are shown for a representative run of DQN and Rainbow. Both networks seem to learn qualitatively similar features, with images featuring vertical lines reminiscent of patterns and smaller objects reminiscent of balls. Further exploration is needed to ground out these evocative appearances.}
  \label{fig:pong_final_layer}
\end{figure}

Such visualizations can also reveal that critical features are being attended to (figure \ref{fig:freeway_first_fc} and  supplemental figure \ref{fig:pong_final_layer}).
Overall, these visualizations demonstrate the potential of this kind of technique, and we believe that many useful further insights may result from a more systematic application and investigation of this and many of the other interesting visualization techniques implemented by Lucid, which can now easily be applied to Atari Zoo models. Also promising would be to further explore regularization to constrain the space of synthetic inputs, e.g.\ a generative model of Atari frames in the spirit of \citet{nguyen2017plug} or similar works.


\vspace{-0.05in}
\section{Discussion and Conclusions}


There are many follow-up extensions that the initial explorations of the zoo raise. One natural
extension is to include more DRL algorithms (e.g.\ TRPO or PPO \citep{schulman2017proximal}).
Beyond algorithms, there are many alternate architectures that might have interesting effects
on representation and decision-making, for example recurrent architectures, or architectures that exploit attention. 
Also intriguing is examining the effect of the
incentive driving search:
Do auxiliary or substitute objectives qualitatively change DRL representations, e.g.\ as in UNREAL 
\citep{jaderberg2016reinforcement}, 
curiosity-driven exploration \citep{pathak2017curiosity}, 
or novelty search \citep{conti:improving}? 
How do the representations and features of meta-learning agents such as 
MAML \citep{finn2017model} change as they learn a new task?
Finally, there are other analysis tools that could be implemented, which might illuminate other interesting properties of DRL algorithms and learned representation, e.g.\ the image perturbation analysis of \citet{greydanus:visualizing} or a variety of sophisticated neuron visualization techniques \citep{nguyen2017plug}.
We welcome community contributions for these algorithms, models, architectures, incentives, and tools.


While the main motivation for the zoo was to reduce friction for research into understanding and visualizing 
the behavior of DRL agents, it can also serve as a platform for other research questions. For example, having
a zoo of agents trained on individual games, for different amounts of data, also would reduce friction for
exploring transfer learning within Atari, 
i.e.\ whether experience learned on one game can quickly benefit on another game. 
Also, by providing a huge library of
cached rollouts for agents across algorithms, the zoo may be interesting in the context of learning from demonstrations, 
or for creating generative models of games. 
In conclusion, we look forward to seeing how this dataset will be used by the community at large.

\pagebreak
\bibliographystyle{named}
\bibliography{cites}

\clearpage
\onecolumn
\beginsupplement

\section*{Supplementary Material}

The following sections contain additional figures, and describe in more detail the experimental setups
applied in the paper's experiments.

\section{Quantitative Analysis Details}

\begin{figure}
\begin{center}
\begin{subfigure}{.8\textwidth}
  \centering
  \includegraphics[width=.8\linewidth]{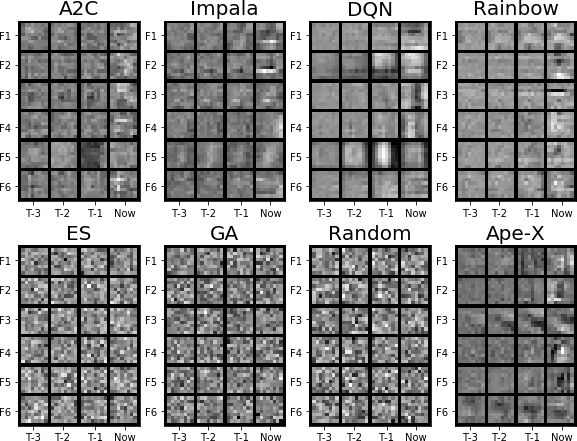}
  \caption{Seaquest}
\end{subfigure}  \\
	\vspace{0.2in}
\begin{subfigure}{.8\textwidth}
  \centering
  \includegraphics[width=.8\linewidth]{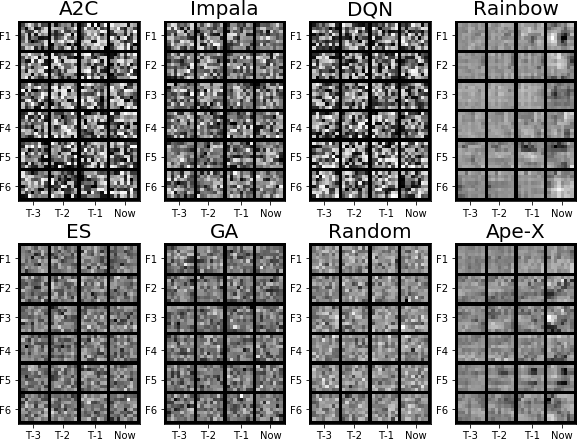}
  \caption{Venture}
\end{subfigure}
\end{center}

	\caption{\textbf{Learned Convolutional Filters}. Shown are first-layer convolutional filters taken from DNNs trained by each algorithm, as well as random filters drawn from a normal distribution (Random). In games in which they exceed random performance, filters for the gradient-based algorithms often have spatial structure and sometimes resemble edge detectors, and the intensity of weights often degrades into the past (i.e.\ the left-most patches). This can be seen for all gradient-based methods in (a) Seaquest; when gradient-based methods fail to learn, as DQN and A2C often do in (b) Venture, their filters then appear more random (this effect is consistent across runs). Filters for the evolutionary algorithms appear less regular, even when their performance is competitive with the gradient-based methods.}
\label{fig:convfilters}
\end{figure}

\begin{figure}
\begin{center}
  \centering
  \includegraphics[width=.9\linewidth]{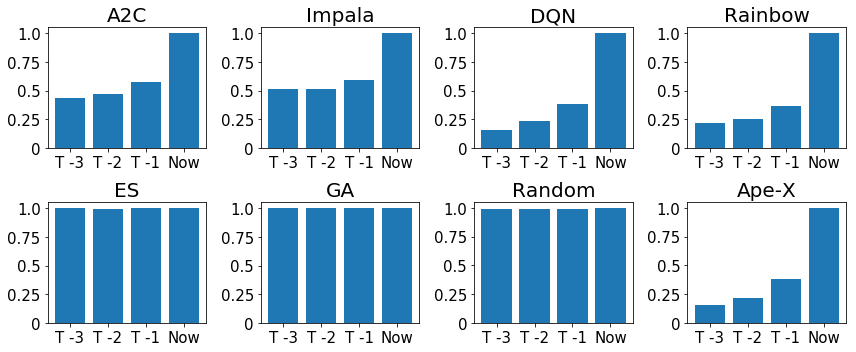}
	\caption{\textbf{Significance of Time Across Models}. Filter weight magnitudes across input patches are shown averaged across a representative sample of ALE games with $3$ independent runs each for each DRL algorithm; also included for analysis are random filters drawn from a normal distribution (Random). Before averaging, past-frame weight magnitudes are normalized by that of weights attending to the most recent observation (i.e.\ the present magnitudes are anchored to $1.0$). For the gradient-based DRL algorithms (Ape-X, Rainbow, DQN, \& A2C), filter weights are stronger when connected to the current frame than to historical frames. Interestingly, such a trend is not seen for the evolutionary algorithms; note that ES includes L2 regularization, so this effect is not merely an artifact of weight decay being present in the gradient-based methods only. The effect is also present when looking at individual games (data not shown).}
\label{fig:temporal}
\end{center}
\end{figure}

Figure \ref{fig:convfilters} shows a sampling of first-layer convolutional filters from final trained models, and figure \ref{fig:temporal} highlights that such filters often differentially attend to the present over the past. 

\subsection{Further Study of Temporal Bias in DQN}
\label{sec:memory}

\begin{figure}
\begin{center}
\includegraphics[width=1.0\linewidth]{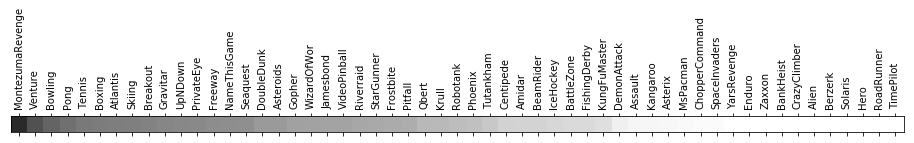}
\end{center}
\caption{\textbf{Attention to the past in DQN}. DQN's tendency to focus on the present relative to the past (as measured by filter magnitudes from different input frames), is shown across 55 ALE games. From left to right, the amount of present-bias increases, e.g.\ the games at the left seemingly have greater use for information stored in the past three frames relative to the games on the right.}
\label{fig:dqn_attention}
\end{figure}

As an exploration of the connection between the information theoretic measure of memory-dependent action in \citetNew{dann2016memory} and the pattern highlighted in this paper (i.e.\ the strength
of filter weights in the first layer of convolutions may highlight a network's reliance on the past), we examined first-layer filters in DQN across all 55 games. A simple metric of present-focus is
the ratio of average weight magnitudes for the past three frames to the present frame. When sorted by this metric (see figure \ref{fig:dqn_attention}), there is high agreement with the 8 games
identified by \citetNew{dann2016memory} that use memory. In particular, three out of the top four games identified by our metric align with theirs; as do six out of the top twelve games, considered
among the games that overlap between their 49 and our 55.

\subsection{Observation and Parameter Noise Details}
\label{sec:parameter_noise}


\begin{figure}
\begin{center}
\begin{subfigure}{.4\textwidth}
  \centering
  \includegraphics[width=0.9\linewidth]{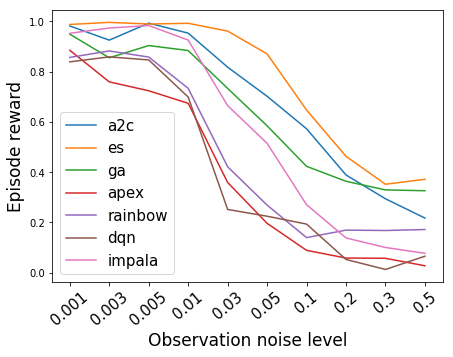}
  \caption{Normalized by algorithm best}
\end{subfigure}%
\begin{subfigure}{.4\textwidth}
  \centering
  \includegraphics[width=0.9\linewidth]{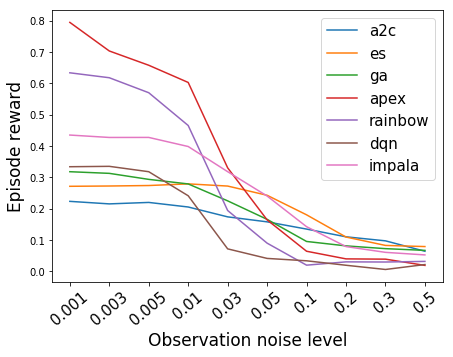}
  \caption{Normalized by best over all algorithms}
\end{subfigure}
	\caption{\textbf{Robustness to Observation Noise}. How performance of trained policies degrades with increasing severe normally-distributed noise is shown, averaged over three independent runs across the analysis subset of games. The figure hows performance degrades (a) relative to baseline performance by that algorithm on each game, and (b) by the best performance of any algorithm on each game. Zero performance in this chart represents random play. The conclusion is that the policy search algorithms show less steep degradation relative to (a) their own best performance; although this is confounded by (b) the overall better absolute performance of the value-based methods. Follow-up analysis will control for performance, by using the Atari Zoo human-performance frozen models.}
\label{fig:obs_noise}
\end{center}
\end{figure}

\begin{figure}
\begin{center}
\begin{subfigure}{.45\textwidth}
  \centering
  \includegraphics[width=.9\linewidth]{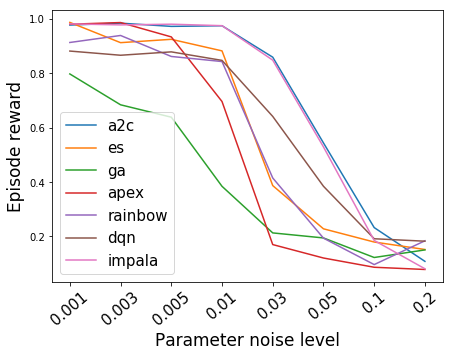}
  \caption{Performance relative to algorithm best}
\end{subfigure}%
\begin{subfigure}{.45\textwidth}
  \centering
  \includegraphics[width=.9\linewidth]{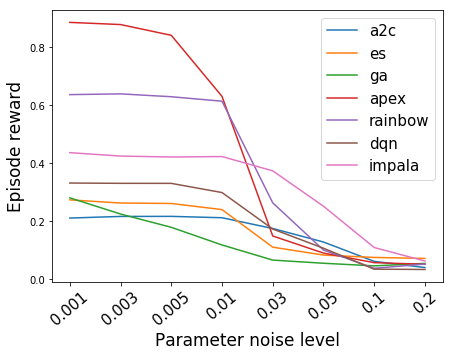}
  \caption{Performance relative to best over algorithms}
\end{subfigure}
\caption{\textbf{Robustness to Parameter Noise}. 
 How performance of trained policies degrades with increasing severe normally-distributed parameter noise is shown, averaged over three independent runs across the analysis subset of games. The figure hows performance degrades (a) relative to baseline performance by that algorithm on each game, and (b) by the best performance of any algorithm on each game. Zero performance in this chart represents random play. Interestingly, the two policy-gradient methods demonstrate a very similar algorithm-best profile that is more robust than the other methods; our prior hypothesis was that the evolutionary algorithms might exhibit higher robustness by this measure (given that they are trained with parameter perturbations).
}
\label{fig:param_noise}
\end{center}
\end{figure}

Figure \ref{fig:obs_noise} shows robustness to observation noise for games in the analysis subset. 
Beyond observation noise, 
another interesting property of learning algorithms is the kind of local optimum they find in the parameter space, i.e.\ whether the learned function is smooth or not in the area of a found solution. One gross tool for examining this property is to test the robustness of policies to parameter perturbations. It is plausible that the evolutionary methods would be more robust in this way, given that they are trained through parameter perturbations. To measure this, we perturb the convolutional weights of each DRL agent with increasingly severe normally-distributed noise. We perturbed the convolutional weights only, because that part of the DNN is identical across agents, whereas the fully-connected layers sometimes vary in structure across DRL algorithms (e.g.\ value-based algorithms like Rainbow or Ape-X that include features that require post-convolutional architectural changes). Figure \ref{fig:param_noise} shows the results across games and algorithms; while no incontrovertible trend exists, the two policy-gradient methods (A2C and Impala) show greater relative robustness than the other methods.

Note that for both algorithm-best performance plots (i.e.\ figures \ref{fig:obs_noise}a and \ref{fig:param_noise}a), three games were excluded from analysis because at least one of the DRL algorithms performed worse than a random policy before pertubation; including them would have conflated performance and robustness, which would undermine the purpose of the plot.

\subsection{Distinctiveness of Policies Learned by Algorithms}
\label{sec:state}

\begin{figure}
 \centering
\includegraphics[width=.50\linewidth]{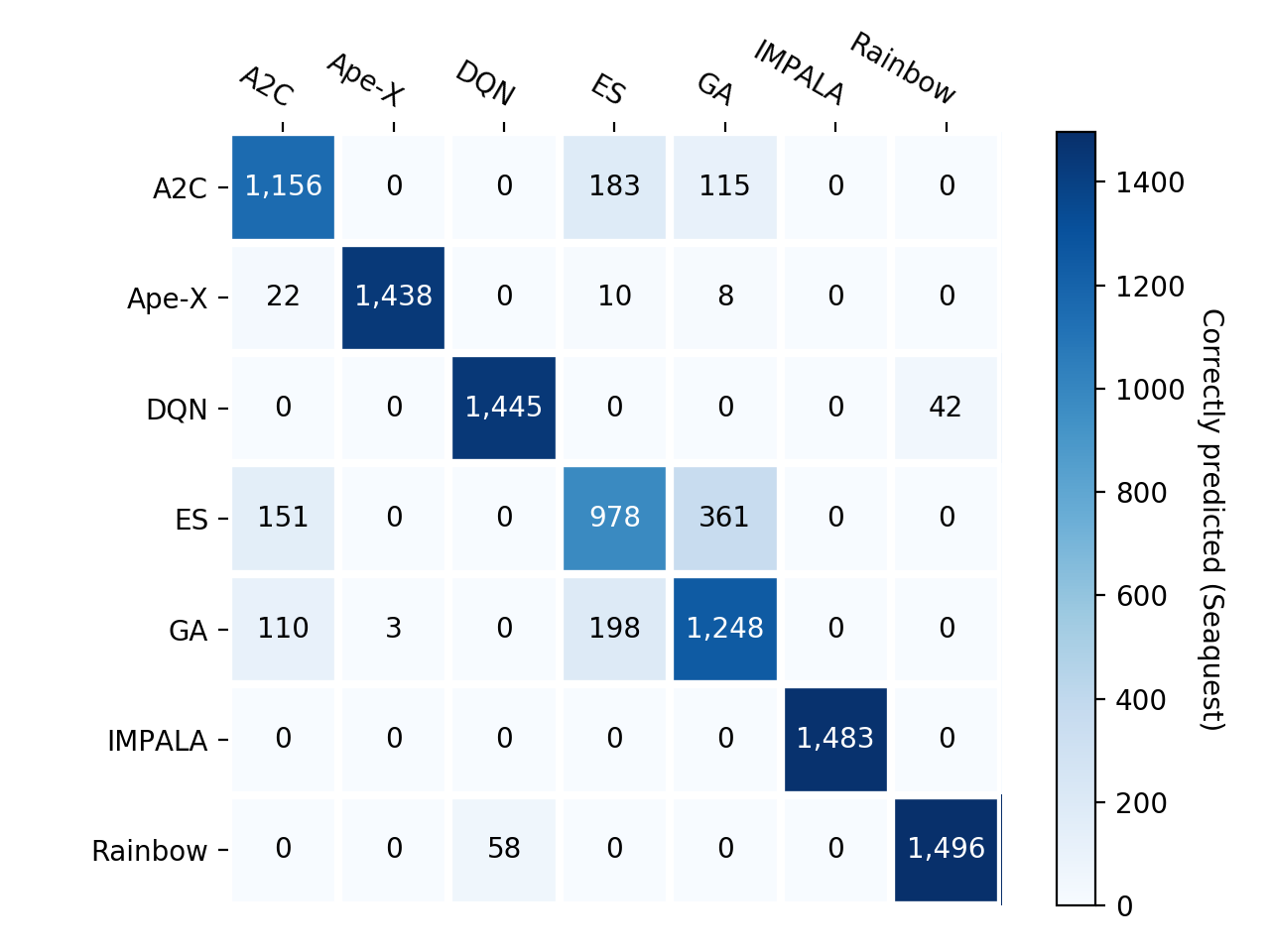}
	\caption{\textbf{Confusion matrix for Seaquest frame classification}.  The cell in the $i$th row and the $j$th column denotes the number of frames generated by the algorithm in row $i$ that are predicted to be generated by the algorithm in column $j$. The conclusion is that in this game, there is a cluster of confusion among many of the direct policy search algorithms (ES, A2C, and GA), highlighting that they are converging to similarly sub-optimal behaviors.}
\label{fig:classify_cf}
\end{figure}

\begin{figure}
 \centering
\includegraphics[width=.55\linewidth]{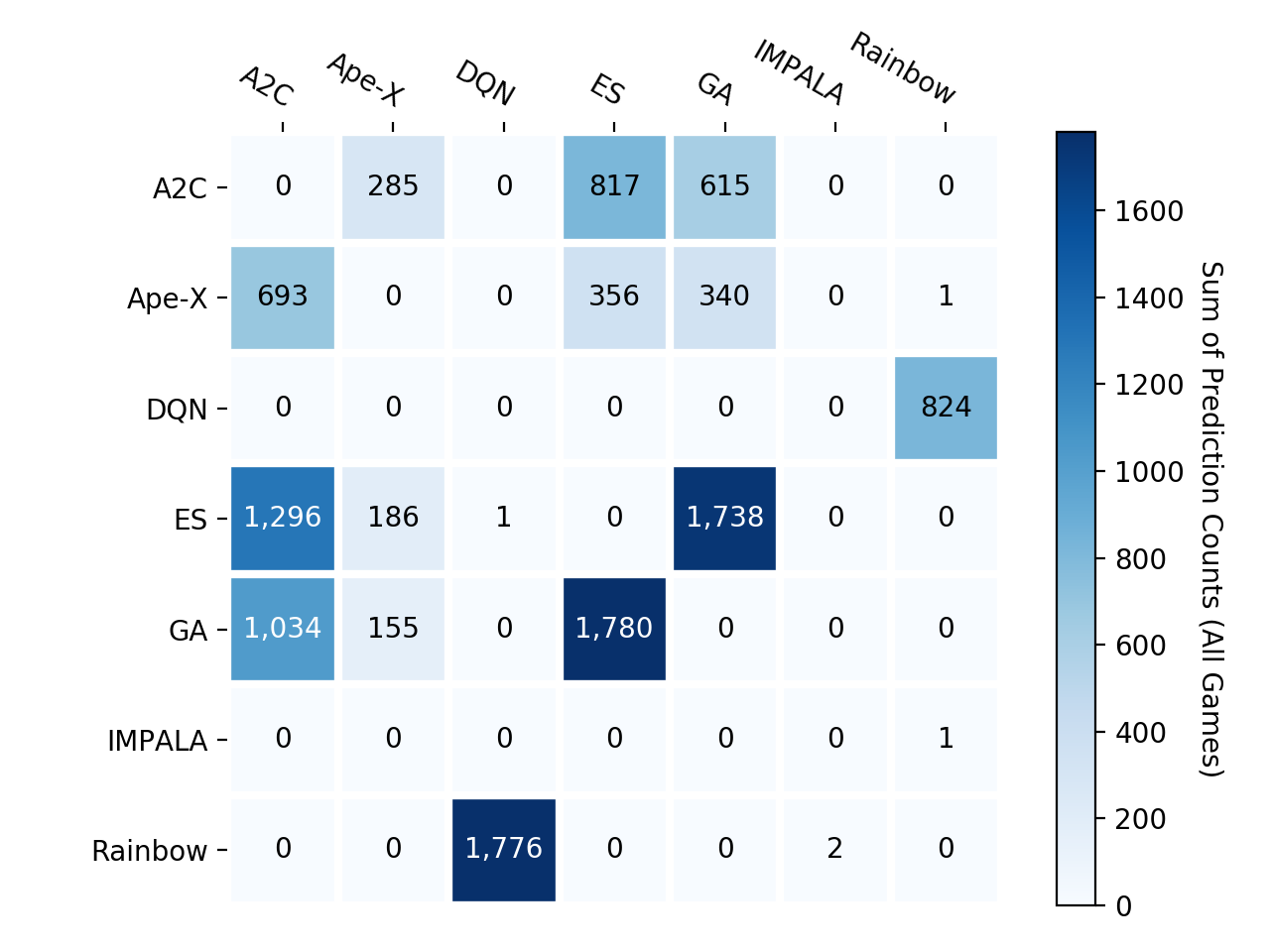}
\caption{\textbf{Confusion matrix summed across all games}. The cell in the $i$th row and the $j$th column denotes the total number of frames from the rollouts of the algorithm in row $i$ predicted to be a from rollouts of the algorithm in column $j$. The true positive predictions are reset to 0 to highlight the false positives. 
	}
\label{fig:classify_cf_all}
\end{figure}

We use only the ``present'' channel of each gray-scale observation frame (i.e.\ without the complete four-frame stack) to train a classifier for each game. The classifier consists of two convolution layers and two fully connected layers, and is trained with early stopping to avoid overfitting. For each game, 2501 frames are collected from multiple evaluations by each model. The reported classification results use 20\% of the frames as test set. Figure \ref{fig:classify_cf} visualizes the confusion matrix for Seaquest frame classification, while figure \ref{fig:classify_cf_all} shows a confusion matrix summed across games. 

\begin{figure}
\centering
\includegraphics[width=.75\linewidth]{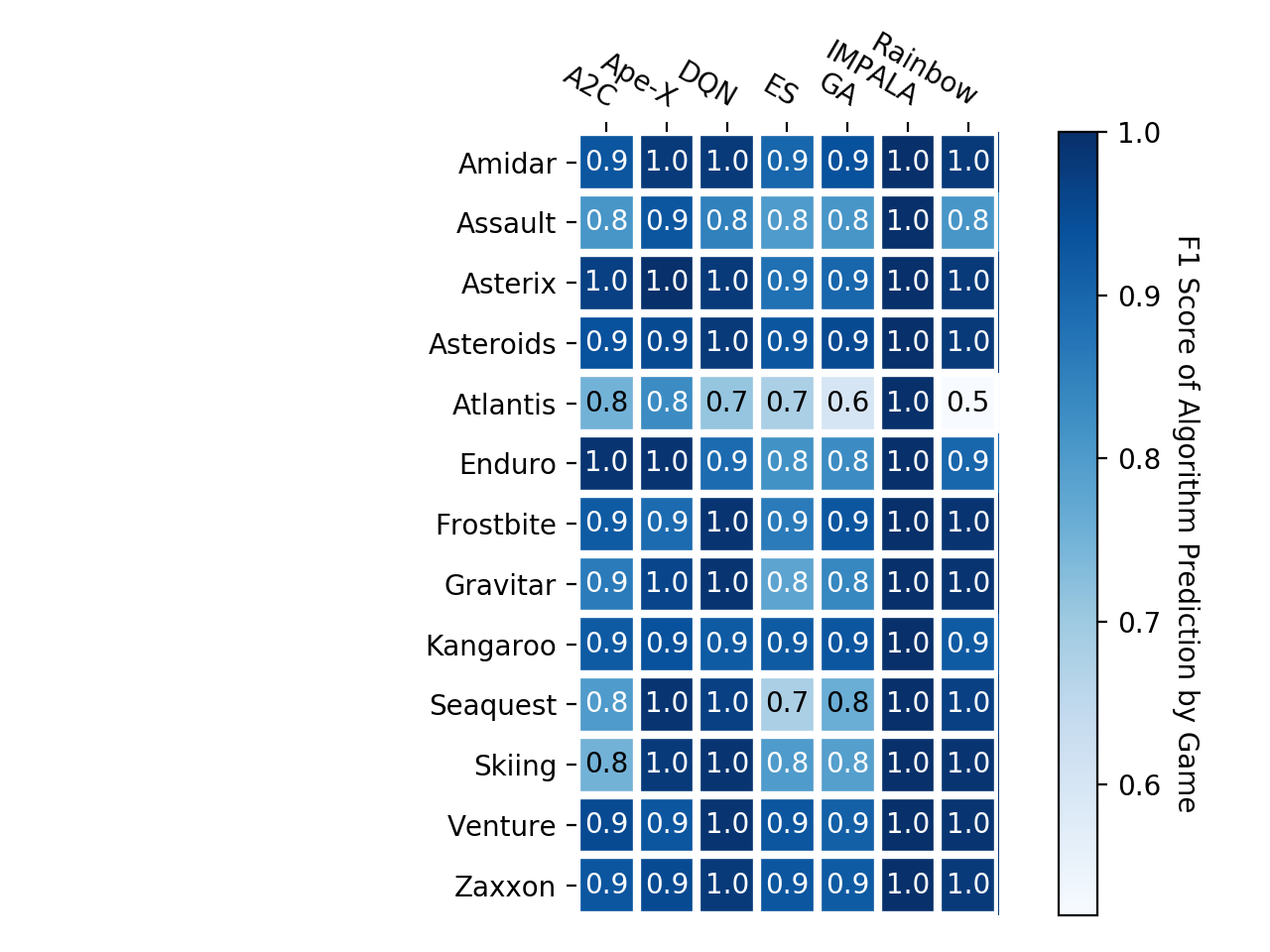}
	\caption{\textbf{F1 scores for frame classification}. F1 score is defined as $2\times\frac{\texttt{precision} \times \texttt{recall}}{\texttt{precision} + \texttt{recall}}$. We observe the classifier distinguishes each algorithm in all environments with at least 0.5 score.} 
\label{fig:f1_algo_by_game}
\end{figure}

We also provide summaries of F1 classification performance: Figure \ref{fig:f1_algo_by_game} summarizes classification performance across DRL algorithms and games, while table \ref{tab:classify_avg_f1} shows performance averaged across games. 

\begin{table}[h]
 \centering
\begin{tabular}{l r}
\toprule
Game & Mean F1 \\
\midrule
Amidar & 0.96 \\
Assault & 0.86 \\
Asterix & 0.96 \\
Asteroids & 0.96 \\
Atlantis & 0.73 \\
Enduro & 0.92 \\
Frostbite & 0.94 \\
Gravitar & 0.92 \\
Kangaroo & 0.93 \\
Seaquest & 0.88 \\
Skiing & 0.9 \\
Venture & 0.96 \\
Zaxxon & 0.96 \\
\bottomrule
\vspace{0.05cm}
\end{tabular}
 \caption{\textbf{Average F1 scores by game.} The score is an unweighted average of F1 scores across all algorithm. Lower scores indicate games for which different DRL algorithms are less distinguishable from each other.}
\label{tab:classify_avg_f1}
\end{table}

\section{Visualization Details}

\begin{figure}
\centering
\includegraphics[width=.85\linewidth]{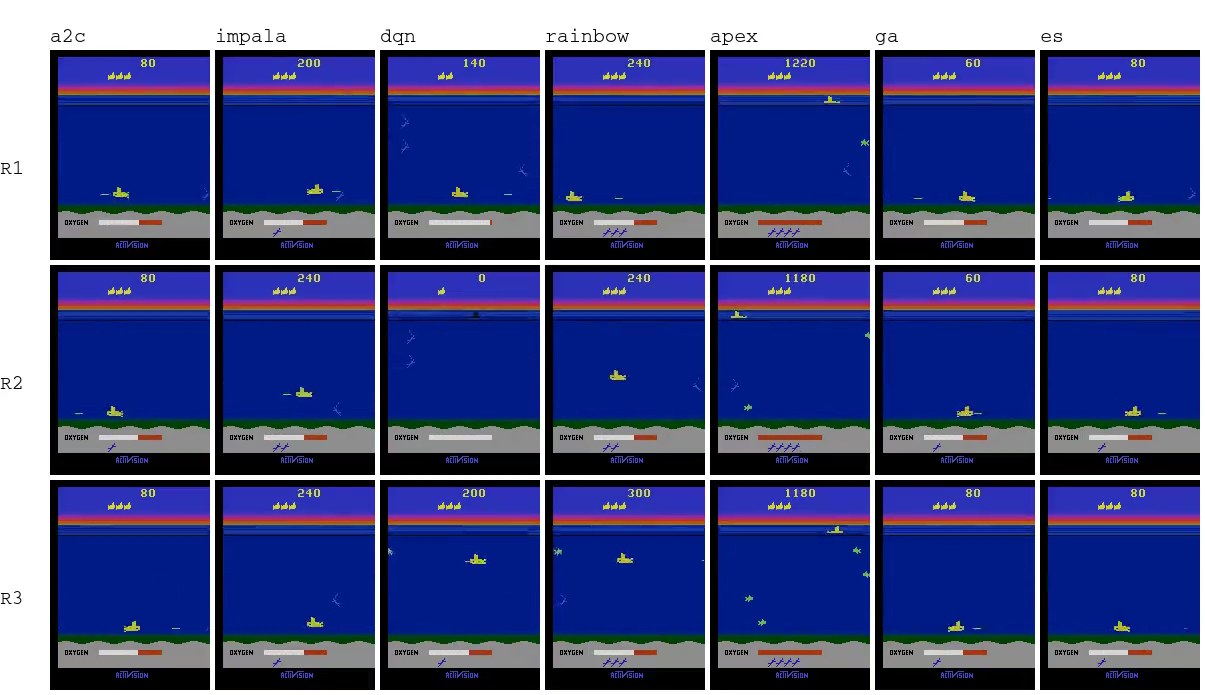}
\caption{\textbf{Grid of Rollout Videos in Seaquest}. The vertical axis represents different independently-trained models, while the horizontal axis represents the DRL algorithms
	included in the Atari Zoo. In Seaquest, one objective is to control a submarine to shoot fish without getting hit by them, and another is to avoid running out of oxygen by intermittently resurfacing. All three independent runs of A2C, GA, and ES converge to the same sub-optimal behavior: They dive to the bottom of the ocean, and shoot fish until they run out of oxygen. The value-function based methods exhibit more sophisticated behavior, highlighting that in this game, greedy policy searches may often converge to sub-optimal solutions, while learning the value of state-action pairs can avoid this pathology. Video is available at:  
	\url{http://bit.ly/2XpD5kO}
	}
\label{fig:grid_rollout}
\end{figure}

This section provides more details and figures for the visualization portion of the paper's analysis.
Figure \ref{fig:grid_rollout} shows one frame of a collage of simultaneous videos that give a quick high-level
comparison of how different algorithms and runs are solving an ALE environment.
Figure \ref{fig:grid_rollout} shows one frame of a video that simultaneously shows a DNN agent acting
in an ALE environment and all of the activations of its DNN.

\begin{figure}
\centering
\includegraphics[width=.9\linewidth]{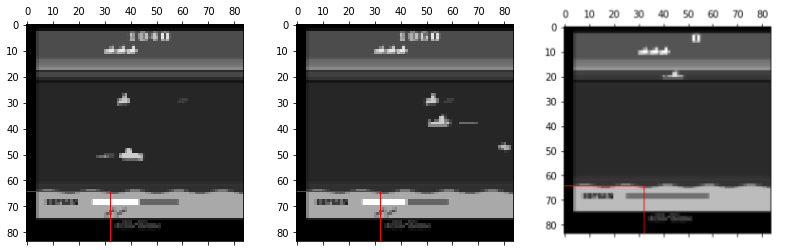}\\
\includegraphics[width=.9\linewidth]{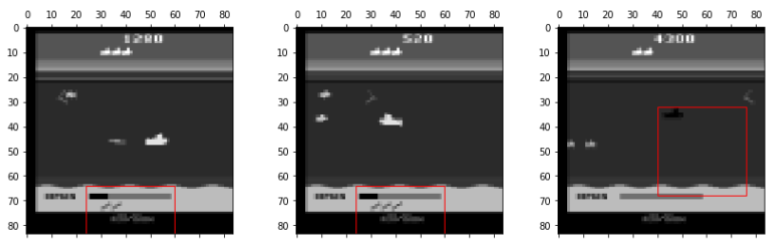}
\caption{\textbf{Location-anchor and oxygen-detector in a Rainbow agent in Seaquest}. The top three images show image patches (red square) that highly-activate a first-layer convolution filter of a Rainbow agent; this filter always activates maximally in the same geometric location, potentially serving as a geometric anchor for localization by down-stream filters. The bottom three images show images patches that highly-activate a separate first-layer filter in the same agent. It detects blinking objects; the submarine can blink before it runs out of oxygen, and the oxygen meter itself blinks when it is running low. 
}
\label{fig:anchors}
\end{figure}

Figure \ref{fig:anchors} shows a second example of how the image-patch finder can help ground out what particular DNN neurons
are learning. 

\subsection{t-SNE Details}
\label{sec:sup_tsne}

To visualize RAM states and high-level DNN representations in 2D, as is typical in t-SNE analysis PCA is first applied to reduce the number of dimensions (to 50), followed by 3000 t-SNE iterations with perplexity of 30. The dimensionality reduction of RAM states is applied across all available runs of DRL algorithms to be jointly embedded. In contrast, dimensionality reduction of high-level DNN representations is particular to a specific model trained by a single DRL algorithm (i.e.\ each run of a DRL algorithm learns its own distinct representation). 

\subsection{Synthetic Input Generation Details}
\label{sec:lucid}

We use the lucid library \citepNew{Lucid} to visualize what types of inputs maximize neuron activations throughout the agents' networks. This study used the trained checkpoints provided by Dopamine \citepNew{bellemare:dopamine} for DQN and Rainbow (although it could be applied to any of the DRL algorithms in the Atari Zoo). These frozen graphs are then loaded as part of a Lucid model and an optimization objective is created. 

An input pattern to the network (consisting of a stack of four 84x84 pixel screens) is optimized to maximize the activations of the desired neurons. Initially, the four 84x84 frames are initialized with random noise. The result of optimization ideally yields visualizations that reveal qualitatively what features the neurons have learned to capture. As recommended in \citetNew{olah2017feature} and \citetNew{mahendran2016visualizing} we apply regularization to produce clearer results; for most images we use only image jitter (i.e.\ randomly offsetting the input image by a few pixels to encourage local translation invariance). For some images, we found it helpful to add total variation regularization (to encourage local smoothness; see \citetNew{mahendran2016visualizing}) and L1 regularization (to encourage pixels that are not contributing to the objective to become zero) on the optimized image.

\section{DRL Algorithm Details and Hyperparameters}
\label{sec:algo_hyper}

This section describes the implementations and hyperparameters used for training the models released with the zoo. The DQN and Rainbow models come from the Dopamine model release \citepNew{bellemare:dopamine}\footnote{The hyperparameters
and training details for Dopamine can be found in https://github.com/google/dopamine/tree/master/baselines/}. The following sections describe the algorithms for the newly-trained models released with this paper.

\subsection{A2C}

The implementation of A2C \citepNew{mnih2016asynchronous} that generated the models in this paper was derived from the OpenAI baselines software package \citepNew{dhariwal:baselines}. It ran with 20 parallel worker threads for 400 million frames; checkpoints occurred every 4 million frames. Hyperparameters are listed in table \ref{tab:hyperparameters_a2c}.

\begin{table}[h]
 \centering
\begin{tabular}{l r r r}
\toprule
\small Hyperparameter & \small Setting \\
\midrule
\small Learning Rate & \small 7e-5 \\
\small $\tau$ & \small 1.0 \\
\small Value Function Loss Coefficient & \small 0.5 \\
\small Entropy Loss Coefficient & \small 0.01 \\
\small Discount factor & \small 0.99 \\
\bottomrule
\vspace{0.05cm}
\end{tabular}
 \caption{\textbf{A2C Hyperparameters.} Population sizes are incremented to account for elites ($+1$). Many of the unusual numbers were found via preliminary hyperparameter searches in other domains.}
\label{tab:hyperparameters_a2c}
\end{table}

\subsection{Ape-X}

The implementation of Ape-X used to generate the models in this paper was based on the one found here: 
\url{https://github.com/uber-research/ape-x}. 
The hyperparameters are reported in Table~\ref{tab:hyperparameters_apex}.

\begin{table}[h]
 \centering
\begin{tabular}{l r r r}
\toprule
\small Hyperparameter & \small Setting \\
\midrule
\small Buffer Size & \small $2 ^ {21}$ \\
\small Number of  Actors & \small 384 \\
\small Batch Size & \small 512 \\
\small n-step & \small 3 \\
\small gamma & \small 0.99 \\
\small gradient clipping & \small 40 \\
\small target network period & \small 2500 \\
\small Prioritized replay $(\alpha, \beta)$ & \small (0.6, 0.4) \\
\small Adam Learning rate & \small 0.00025 / 4 \\
\bottomrule
\vspace{0.05cm}
\end{tabular}
 \caption{\textbf{Ape-X Hyperparameters.} For more details on what these parameters signify, see \protect\citepNew{horgan:distributed}.}
\label{tab:hyperparameters_apex}
\end{table}

\subsection{GA}

The implementation of GA used to generate the models in this paper was based on the one found here: 
\url{https://github.com/uber-research/deep-neuroevolution}. 
The hyperparameters are reported in Table~\ref{tab:hyperparameters_ga} and were found through random search.

\begin{table}[h]
 \centering
\begin{tabular}{l r r r}
\toprule
\small Hyperparameter & \small Setting \\
\midrule
\small $\sigma$ (Mutation Power) & \small 0.002 \\
\small Population Size & \small 1000 \\
\small Truncation Size & \small 20 \\
\bottomrule
\vspace{0.05cm}
\end{tabular}
 \caption{\textbf{GA Hyperparameters.} For more details on what these parameters signify, see \protect\citepNew{such:deep}.}
\label{tab:hyperparameters_ga}
\end{table}

\subsection{ES}
The implementation of ES used to generate the models in this paper was based on the one found here: \url{https://github.com/uber-research/deep-neuroevolution}. The hyperparameters reported in Table~\ref{tab:hyperparameters_es} were found via preliminary search and are similar to those reported in \citepNew{conti:improving}.

\begin{table}[h]
 \centering
\begin{tabular}{l r r r}
\toprule
\small Hyperparameter & \small Setting \\
\midrule
\small $\sigma$ (Mutation Power) & \small 0.02 \\
\small Virtual Batch Size & \small 128 \\
\small Population Size & \small 5000 \\
\small Learning Rate & \small 0.01 \\
\small Optimizer & \small Adam \\ 
\small L2 Regularization Coefficient & \small 0.005 \\
\bottomrule
\vspace{0.05cm}
\end{tabular}
 \caption{\textbf{ES Hyperparameters.} For more details on what these parameters signify, see \protect\citepNew{salimans:es, conti:improving}.}
\label{tab:hyperparameters_es}
\end{table}

\subsection{Impala}

The implementation of Impala used to generate the models in this paper is based on the one found here: \url{https://github.com/deepmind/scalable_agent}. The hyperparameters reported in Table~\ref{tab:hyperparameters_impala} are the same as those reported in \citepNew{espeholt:impala}.

\begin{table}[h]
 \centering
\begin{tabular}{l r r r}
\toprule
\small Hyperparameter & \small Setting \\
\midrule
\small Number of actors & \small 25 \\
\small Image Width & \small 84 \\
\small Image Height & \small 84 \\
\small Grayscaling & \small Yes \\
\small Action Repetitions & \small 4 \\
\small Max-pool over last N action repeat frames & \small 2 \\
\small Frame Stacking & \small 4 \\
\small End of episode when life lost & \small Yes \\
\small Reward Clipping & \small [-1, 1] \\
\small Unroll Length (n) & \small 20 \\
\small Batch size & \small 32 \\
\small Discount ($\gamma$) & \small 0.99 \\
\small Baseline loss scaling & \small 0.5 \\
\small Entropy Regularizer & \small 0.01 \\
\small RMSProp momentum & \small 0.0 \\
\small RMSProp $\epsilon$ & \small 0.01 \\
\small Learning rate & \small 0.0006 \\
\small Clip global gradient norm & \small 40.0 \\
\small Learning rate schedule & \small Anneal linearly to 0 from beginning to end of training \\
\bottomrule
\vspace{0.05cm}
\end{tabular}
 \caption{\textbf{IMPALA Hyperparameters.} For more details on what these parameters signify, see \protect\citepNew{espeholt:impala}.}
\label{tab:hyperparameters_impala}
\end{table}


\bibliographystyleNew{named}
\bibliographyNew{cites}

\end{document}